# Correlation Preserving Sparse Coding Over Multi-level Dictionaries for Image Denoising

# Correlation Preserving Sparse Coding Over Multi-level Dictionaries for Image Denoising

Rui Chen, *Member, IEEE,* Huizhu Jia, Xiaodong Xie and Wen Gao, *Fellow, IEEE*

*Abstract*—In this letter, we propose a novel image denoising method based on correlation preserving sparse coding. Because the instable and unreliable correlations among basis set can limit the performance of the dictionary-driven denoising methods, two effective regularized strategies are employed in the coding process. Specifically, a graph-based regularizer is built for preserving the global similarity correlations, which can adaptively capture both the geometrical structures and discriminative features of textured patches. In particular, edge weights in the graph are obtained by seeking a nonnegative low-rank construction. Besides, a robust locality-constrained coding can automatically preserve not only spatial neighborhood information but also internal consistency present in noisy patches while learning overcomplete dictionary. Experimental results demonstrate that our proposed method achieves state-of-the-art denoising performance in terms of both PSNR and subjective visual quality.

## I. Introduction

THE objective of image denoising task is to recover the clear image from a noisy measurement while preserving its main informative features such as the edges and textures. The noise introduced during the image acquisition process is generally assumed to be an additive zero-mean white and homogeneous Gaussian distribution [1]. The estimation for unknown original image is actually an ill-posed inverse linear problem due to the inadequate constraints [2]. To derive the closed-form solution, certain prior information of the image need be utilized to regularize the recovery process. It is proved that natural signals and images have an essentially sparse representation [3]. Hence, the sparsity presented in analytical transform domain or learned dictionary can be fully exploited in restoration problem.

Many successful denoising methods have been proposed base on the correlations which reliably exist in spatial domain, transform domain or sparse coding for image patches [4]. A pioneer work called as non-local mean method (NLM) is to remove the noise by averaging the pixels with the spatial self-similarity correlation between patches from different locations [5]. According to the non-local and redundant correlations in the image patches, the famous benchmark BM3D [6] groups structurally similar patches to form 3D stack and then performs the collaborative filtering in DCT or Haar wavelet transforms. Recent advances for image denoising rely on coding correlation of image patches which can be sparsely represented by linear combination of basis vectors from an over-complete dictionary [7]-[15]. Representative dictionary learning-based methods are learned simultaneous sparse coding (LSSC) [7], clustering based sparse representations (CSR) [8] and its non-local improved version NCSR [9], K-clustering with singular value decomposition (K-SVD) [10] and its variants [11]. Besides, the combination of sparse models and low rank completion of data matrix is used to denoise the image [16]. Through performance analysis for these state-of-the-art methods, the strong denoising capability roots in the utilization of the multi-level correlations which mainly include the geometric similarity in external patches and the coding consistency of similar features in internal patches [4]. However, the current denoising methods based on dictionary learning have some limitations to further improve the results because the similarity correlations of image patches may be damaged in sparse coding process. Specifically, high-dimension data are converted to vector format [17], hence this may lose high local correlations among the neighboring pixels. Moreover, small patch size can decrease the complexity of textures [18], and thus the correlations of non-local structural similarity will be affected. In parallel, when the dictionary is adaptively learned from the image itself using the patches containing the high noise levels, the correlations of nonzero coefficients will not correctly reflect the stable properties of feature space, which results in amount of artifacts. Although some later works [16],[19] have made efforts to tackle these problems, the optimal denoising gain is still not achieved.

In this letter, we address these issues and propose a novel denoising method, which makes a full exploitation of structure and feature subspaces in the image. Leveraging on the low-rank representation (LRR) model, we have designed an adaptive graph Laplacian (AGL) model as the regularizer to preserve the global similarity correlations and enhance the dissimilarity with noises. Besides, the locality constraint (LC) coding is used to preserve the consistence in internal basis sets. By incorporating these preserving terms, the noise can be effectively removed.

## II. Correlation Preserving Sparse Coding

In this section, we introduce the sparse coding scheme with the properties of AGL and LC. The details of two correlation preserving strategies are described below.

### A. AGL Regularized Coding

The AGL-based sparse coding scheme can explicitly take into account the manifold structure of the image data. Based on

This work was supported by NSFC under Grant 61171139 and 61520106004, Major National Scientific Instrument and Equipment Development Project of China under Grant 2013YQ030967, and China Postdoctoral Science Foundation under Grant 2014M560852.

The authors are with National Engineering Laboratory for Video Technology, Peking University, Beijing 100871, China (e-mail: {chenrui2014, hzjia, donxie, wengao}@pku.edu.cn).

the spectral graph theory, AGL is adopted as a smooth operator to preserve the manifold structure. The obtained representations vary smoothly along the geodesics of the data manifold. Then the basis vector can encode the intrinsic structure embedded in new space according to the manifold assumption [20]. Given a set of data point $\mathbf{Y} \in \mathbb{R}^{d \times n}$, we can construct a nearest neighbor graph $G$ with $n$ vertices denoting all data points. Let $\mathbf{W} \in \mathbb{R}^{n \times n}$ be a weight matrix of $G$. The similarity can be measured by the edge weight and thus the affinity between the vertices is preserved. That is, the geometrical structure of the data space can be described by AGL. Mathematically, this relationship is formulated by the following regularization function:

$$R(z) = \frac{1}{2} \sum_{i=1}^{n} \sum_{j=1}^{n} (z_i - z_j) \mathbf{W}_{ij} = \mathbf{Tr}(\mathbf{ZLZ}^T). \quad (1)$$

where $z_i$ and $z_j$ are the mappings of two points $y_i$ and $y_j$ under some transform, respectively. $\mathbf{L} = \Lambda - \mathbf{W}$ is the Laplacian matrix. The degree matrix $\Lambda$ is a diagonal matrix whose diagonal elements are the sum of the elements $\{w_{ij}\}$ in weight matrix.

To take advantage of the desired spectral properties, the matrix $\mathbf{L}$ is further normalized. Applying the fast symmetry preserving matrix balancing procedure [21] to $\mathbf{W}$ yields the doubly stochastic filtering matrix $\mathbf{K}$ and returns a diagonal scaling matrix $\mathbf{C}$. Then the normalized matrix $\bar{\mathbf{L}}$ is defined as

$$\bar{\mathbf{L}} = \mathbf{I} - \mathbf{K} = \mathbf{I} - \mathbf{C}^{-1/2} \mathbf{W} \mathbf{C}^{-1/2}. \quad (2)$$

The matrix $\bar{\mathbf{L}}$ is symmetric and positive semi-definite, which can be directly interpreted as a data-adaptive Laplacian filter with the expected behavior. Compared with the un-normalized graph Laplacian, $\bar{\mathbf{L}}$ can provide better performance.

*B. Edge Weight Computation Using LRR*

The LRR model is based on the assumption that data are approximately sampled from low-dimensional subspaces. For a set of data samples, LRR finds the lowest rank representation of all data. It has been shown that LRR is efficient in exploring low-dimensional subspace structures embedded in data. Let the matrix $\mathbf{Y} \in \mathbb{R}^{d \times n}$ be sampled from independent subspaces. Then each column can be represented by linear combination of bases in the dictionary $\mathbf{A}$. By imposing the most sparsity and lowest rank constraints, the coefficient matrix $\mathbf{Z}$ can be reconstructed by solving the following optimization problem

$$\begin{cases} \min_{Z,E} \|\mathbf{Z}\|_* + \beta \|\mathbf{Z}\|_1 + \alpha \|\epsilon\|_1 \\ s.t. \ \mathbf{Y} = \mathbf{AZ} + \epsilon, \ \mathbf{Z} \geq 0 \end{cases}. \quad (3)$$

where $\|\mathbf{Z}\|_*$ is the nuclear norm defined as the sum of all singular values of $\mathbf{Z}$, and $\|\epsilon\|_1$ is the $l_1$-norm of noise term. $\beta > 0$ is a parameter to balance between the low-rankness and sparsity. The parameter $\alpha > 0$ is used to balance the effect of noise, which is set empirically. The LRR can extract the global structures of data $\mathbf{Y}$, while the sparsity can capture the local relevance of each data vector. The problem (3) can be solved by the fast method adopted in [22]. This alternating direction method uses less auxiliary variables and no matrix inversions, and hence it can convergence fast to the minimum solution.

Given an appropriately designed dictionary $\mathbf{A}$, the optimal solution $\mathbf{Z}^*$ of low-rank recovery can accurately reveal some underlying correlations of data. The $ij$-th element of $\mathbf{Z}^*$ reflects the similarity between samples $y_i$ and $y_j$. The sparse constraint ensures that the graph derived from $\mathbf{Z}^*$ is sparse. The low-rank guarantees that the coefficients of samples are highly correlated in the same subspace, so $\mathbf{Z}^*$ can capture the global structures of the whole data. Based on these low-rank characteristics, after obtaining the optimal coefficient matrix $\mathbf{Z}^*$, the graph weight $\mathbf{W}$ is derived from it. The matrix $\mathbf{W}$ is defined as follows:

$$\mathbf{W} = (\mathbf{Z}^* + (\mathbf{Z}^*)^T)/2. \quad (4)$$

In practice, for preventing noises to affect the dependencies of graph adjacent structures [22], the coefficients in the weight matrix are set zeros under the given threshold $T$. Let the parameter $\sigma$ be stand deviation of Gaussian noise. The region size is set as $l \times l$. The value of $T$ is computed as follows:

$$T = \sigma \sqrt{2 \log l^2}. \quad (5)$$

*C. LC-based Coding*

This coding scheme utilizes the LC to project local features into the manifold space where their geometric structure can be more easily identified. Hence, the learned dictionary can best reconstruct images while preserving the locality correlations. Let $\mathbf{F} = [f_1, f_2, ..., f_N]$ denote the local descriptors of an image, which can be converted into a set of codes $\mathbf{E} = [e_1, e_2, ..., e_N]$. Given the trained dictionary $\mathbf{B}$, the locality-constrained coding can find the coefficient $e_i$ for each feature $f_i \in \mathbb{R}^k$ by minimizing the objective function, which is given by

$$\begin{cases} \min_{E} \sum_{i=1}^{N} \|f_i - \mathbf{B} e_i\|_2^2 + \gamma \|\varphi_i \odot e_i\|_2^2 \\ s.t. \quad \mathbf{1}^T e_i = 1, \quad \forall i \end{cases}. \quad (6)$$

where the symbol $\odot$ denotes the element-wise multiplication. $\varphi_i \in \mathbb{R}^M$ is the locality adaptor that weights each basis vector proportional to its similarity to the input descriptor $f_i$. Finally, the locality adaptor $\varphi_i$ is formulated as

$$\varphi_i = \exp(\frac{dist(f_i, \mathbf{B})}{\delta}). \quad (7)$$

where $dist(f_i, \mathbf{B}) = [dist(f_i, \mathbf{b}_1), ..., dist(f_i, \mathbf{b}_M)]^T$, and $dist(f_i, \mathbf{b}_j)$ is the Euclidean distance between $f_i$ and $\mathbf{b}_j$. Here $\delta$ is used for adjusting the weight of decay speed for the locality adaptor.

The constrain $\mathbf{1}^T e_i = 1$ ensures shift-invariant requirement.

The LC-based coding has several favorable properties that can achieve less reconstruction error in problem (6). It has been suggested that locality is more important than sparsity because locality must lead to sparsity but not necessary vice versa [23]. Due to the over-completeness of dictionary, the coding process might select quite different bases for similar patches to favor sparsity, thus losing correlations between codes. On the other hand, the explicit locality adaptor by $l_2$-norm can ensure that similar patches will have similar codes [24].

### III. PROPOSED DENOISING METHOD

#### A. Denoising Formulation

Consider that the observed image $\mathbf{Y}$ is corrupted by the white Gaussian noise $\mathbf{v}$ with distribution $N(0, \sigma^2)$. This degradation process can be modeled by $\mathbf{Y} = \mathbf{X} + \mathbf{v}$. The clear image vector $\mathbf{X} \in \mathbb{R}^N$ is estimated by using so-called correlation preserving sparse coding (CPSC) model. Note that the AGL regularizer and the LC are incorporated for constructing the CPSC model by means of the complementary combination. Through seeking for the optimal sparse representation of image data different from non-sparse noises, the denoised image $\mathbf{X}$ can be obtained. Based on the unified regularization, our CPSC model for the whole-image denoising problem can be formulated as

$$(\mathbf{X}, \mathbf{D}, \mathbf{S}) = \arg\min_{\mathbf{X},\mathbf{D},\mathbf{S}} \left\{ \begin{array}{l} \|\mathbf{Y} - \mathbf{X}\|_2^2 + \mu \sum_{i=1}^M \|\mathbf{D}s_i - \mathbf{U}_i \mathbf{X}\|_2^2 \\ + \rho \mathbf{Tr}(\mathbf{S}\overline{\mathbf{L}}\mathbf{S}^T) + \lambda \sum_{i=1}^M \|\varphi_i \odot s_i\|_2^2 \end{array} \right\} \quad (8)$$

where the matrix $\mathbf{U}_i \in \mathbb{R}^{r \times N}$ is used to extract the $i$-th patch from the image. Considering both the computational complexity and the utilization of repetition pattern, the image is divide into fully overlapping small patches to deal with. The operator $\mathbf{U}_i \mathbf{X}$ denotes an image patch of size $\sqrt{r} \times \sqrt{r}$ pixels extracted from the clear image $\mathbf{X}$ at location $i$. $\mathbf{S} = [s_1, s_2, ..., s_M] \in \mathbb{R}^{k \times M}$ is the set of sparse coefficient. The estimate of image patch can be sparsely represented by a linear combination of the spare coefficient $s_i \in \mathbb{R}^k$ over the trained dictionary $\mathbf{D} \in \mathbb{R}^{r \times k}$. The regularization parameters $\mu > 0$, $\rho > 0$ and $\lambda > 0$ can be tuned to empirically control the constrained weights, respectively. The matrix $\overline{\mathbf{L}}$ is computed by Eq. (2). According to Eq. (7), the weight vector $\varphi_i$ is obtained. The regularization terms in Eq. (8) can ensure the convergence to a globally optimal solution.

#### B. Optimization for CPSC

With the alternate iteration strategy, the optimization for the CPSC model can be divided into two main stages: updating the dictionary $\mathbf{D}$ stage while fixing the $\mathbf{S}$; and updating the sparse coefficient $\mathbf{S}$ stage whiling fixing the $\mathbf{D}$; until convergence. Then, we obtain the desired clear image $\mathbf{X}$.

**Algorithm 1** The Proposed CPSC Denoising Algorithm.

**Input:** The noisy image $\mathbf{Y}$; Maximum number of iterations $J$; The regularization parameters $\mu$, $\rho$, $\lambda$; The dictionary size $r$ and $k$; The weight parameters $\alpha$, $\beta$, $\delta$.
**Output:** The optimal solutions ($\mathbf{X}$, $\mathbf{D}$, $\mathbf{S}$).
**Initialization:** $J = 100$; $\mu = 1.2$; $\rho = 0.5$; $\lambda = 0.3$; $r = 64$; $k = 256$; $\alpha = 10$; $\beta = 0.2$; $\delta = 80$; $\mathbf{X} = \mathbf{Y}$.
1: **Set** $\mathbf{D}$ as overcomplete DCT dictionary; **Compute** the mean intensity of each patch; **Subtract** the mean intensity;
2: **Obtain** the optimal $\mathbf{Z}^*$ using the LADMAP method in [22]; **Compute** the weight matrix $\mathbf{W}$.
3: **Compute** the filtering matrix $\mathbf{K}$ by applying the fast procedure [21] to $\mathbf{W}$; **Normalize** the matrix $\mathbf{L}$ to obtain $\overline{\mathbf{L}}$.
4: **Repeat** times:
5: **Update** the dictionary $\mathbf{D}$ stage: perform K-SVD scheme.
6: **Update** the coefficient $\mathbf{S}$ stage: search for optimal sparse coefficients using the fast feature-sign procedure.
7: **End** the loop when it satisfies the stopping condition.
8: **Compute** the clear image $\mathbf{X}$.

When $\mathbf{S}$ is assumed known, the dictionary atom $d_i$ is updated by solving the problem (9). The K-SVD algorithm in [11] can be directly used to train and update the dictionary.

$$\mathbf{D} = \arg\min_{\mathbf{D}} \left\{ \mu \sum_{i=1}^M \|\mathbf{D}s_i - \mathbf{U}_i\mathbf{X}\|_2^2 \right\} \; s.t. \; \|d_i\|_2^2 \leq 1. \quad (9)$$

The problem (10) is convex by fixing the dictionary $\mathbf{D}$. Thus, the global minimum can be achieved. We update each vector $s_i$ individually, while holding other vectors $\{s_j\}_{j \neq i}$ constant. The problem (10) can be solved by following the feature-sign search algorithm in [20]. It updates the solution, and linearly searches for the optimal sparse coefficients. The new $s_i$ is further normalized by shift-invariant constraint $\mathbf{1}^T s_i = 1$.

$$\mathbf{S} = \arg\min_{\mathbf{S}} \left\{ \begin{array}{l} \mu \sum_{i=1}^M \|\mathbf{D}s_i - \mathbf{U}_i\mathbf{X}\|_2^2 + \rho \sum_{i,j=1}^M \overline{L}_{ij} s_i^T s_j \\ + \lambda \sum_{i=1}^M \|\varphi_i \odot s_i\|_2^2 \end{array} \right\} \quad (10)$$

Given all sparse codes and the dictionary, a closed-form solution has the quadratic expression. The globally denoised image can be obtained as the following form:

$$\mathbf{X} = (\mathbf{I} + \mu \sum_{i=1}^M \mathbf{U}_i^T \mathbf{U}_i)^{-1} (\mathbf{Y} + \mu \sum_{i=1}^M \mathbf{U}_i^T \mathbf{D}s_i). \quad (11)$$

where $\mathbf{I}$ is the identity matrix. The matrix $\mathbf{U}_i^T$ returns a clean patch to its original location. Finally, the detailed algorithm procedure of CPSC denoising is summarized in **Algorithm 1**.

## IV. EXPERIMENTAL RESULTS

In this section, we have evaluated the denoising performance of the proposed CPSC method. Moreover, we fully compared our method with other three state-of-the-art denoising methods, including BM3D [6], EPLL [12], NSCR [9]. In the experiments, seven natural images are carefully selected for the testing, and their noisy versions are simulated by adding the independent white Gaussian noise with varying deviation level $20 \leq \sigma \leq 80$. For all the dictionary learning methods, we extract image patches of the same size 8×8 with a sliding distance of 1, and randomly select image patches from the training set. As for other main parameters of our method, we set them following the initial values illustrated in **Algorithm 1**.

The image Barbara and Boat are corrupted by the Gaussian noises with the stand deviations of 20 and 60, respectively. As shown in Fig. 1, our method can remove almost all noises and achieve the best subjective results among four methods. Due to the similarity correlation preservation of AGL regularization, the informative structures in the denoised image are very close to the original image while generating few visual artifacts. For the high noise level, it can be clearly seen from Fig. 2 that our method contains both more sharper edges and smoother regions. By encoding the LC for local feature preservation, our method can reliably exclude noises from the candidate patches where strong classifying capability ensures the elegant visual effect.

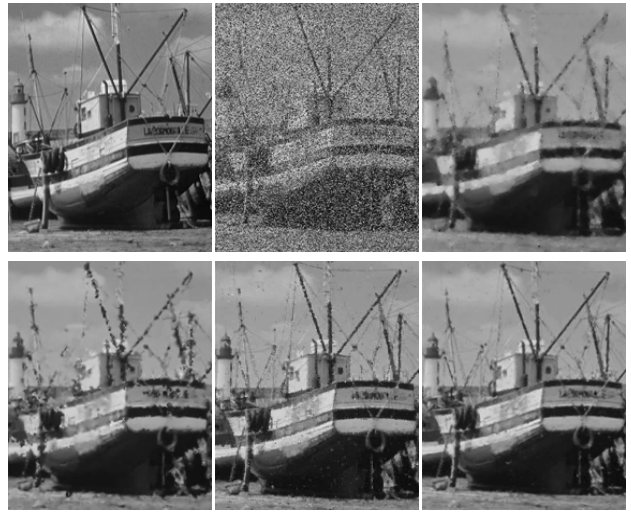

Fig. 2. Denoising performance comparison on the image ***Boat***: original image (top-left); noisy image (top-middle); BM3D (top-right); EPLL (bottom-left); NSCR (bottom-middle); proposed (bottom-right).

To evaluate the objective quality of the denoised images, the values of peak signal-to-noise (PSNR) are computed. The results are shown in Table I. For each image and at each noise level, the highest PSNRs are highlighted in bold. It has shown that our proposed method invariably performs best on all test images. Furthermore, the average statistics for the denoised performance testified that CPSC model can preserve intrinsic correlations well for the various textures, and is capable of discriminating with noises. Meanwhile, our method efficiently convergences to the optimal solution within 100 iterations.

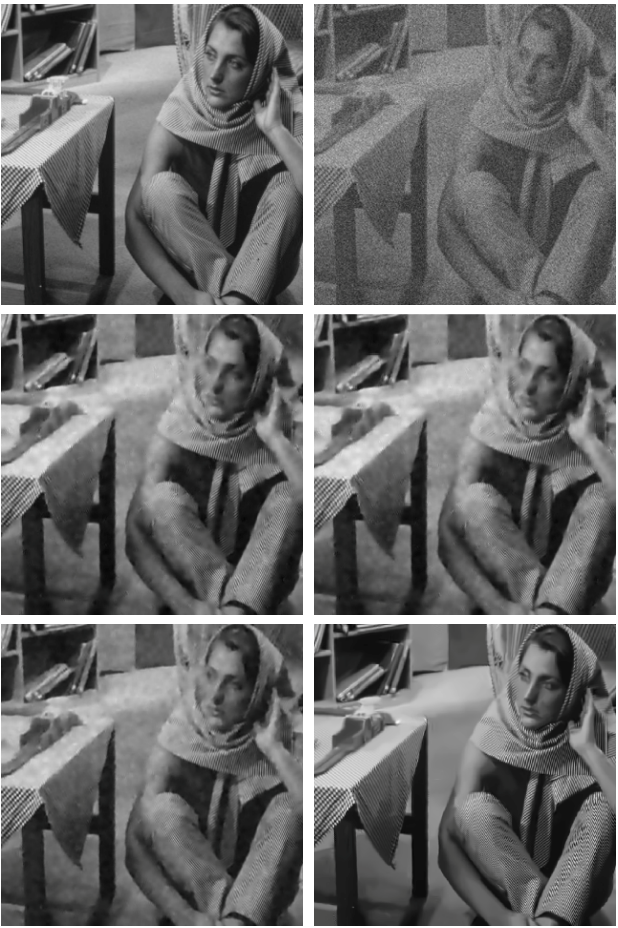

Fig. 1. Denoising performance comparison on the image ***Barbara***: original image (top-left); noisy image (top-middle); BM3D (middle-right); EPLL (middle-left); NSCR (bottom-middle); proposed (bottom-right).

TABLE I
PSNR (dB) RESULTS BY DIFFERENT DENOISING METHODS

| Image | BM3D | EPLL | NSCR | Proposed |
|---|---|---|---|---|
| Monarch | 30.58 | 30.31 | 31.18 | **32.08** |
| Barbara | 29.28 | 29.45 | 29.87 | **31.04** |
| Wheel | 25.86 | 25.12 | 26.32 | **27.54** |
| Airplane | 27.73 | 28.54 | 27.97 | **29.36** |
| Lena | 28.89 | 29.15 | 29.50 | **30.61** |
| House | 26.62 | 26.99 | 27.32 | **27.89** |
| Boat | 28.53 | 28.67 | 29.38 | **29.65** |
| Average | 28.21 | 28.32 | 28.79 | **29.74** |

## V. CONCLUSIONS

In this letter, we have presented an effective and efficient noise removal method based on the unified CPSC model. We utilize the structure space and the mapped feature space as two complementary strategies to address such an ill-posed inverse problem. By integrating the LRR information into the AGL model, global similarity correlation is exploited and encoded better to separate original image data from the noise. Then the LC coding is applied to further preserve the feature correlations over learned codes to remove the residual noise. Experimental results show that the performance of our proposed method is quite competitive with state-of-the-art denoising methods, even much better for the strong noises.

## REFERENCES

[1] H. Liu, R. Xiong, J. Zhang, and W. Gao, "Image denoising via adaptive soft-thresholding based on non-local sanples," in *Proc. IEEE Conf. Computer Vision and Pattern Recognition (CVPR)*, pp. 484-492, 2015.

[2] W. S. Dong, G. M. Shi, Y. Ma, and X. Li, "Image restoration via simultaneous sparse coding: where structured sparsity meets gaussian scale mixture," *Int. J. Comput. Vis.*, vol. 114, no. 2, pp. 217-232, Sep. 2015.

[3] J. Zhang, D. B. Zhao, and W. Gao, "Group-based sparse representation for image restoration," *IEEE Trans. Image Process.*, vol. 23, no. 8, pp. 3336-3351, Nov. 2014.

[4] L. Shao, R. M. Yan, X. L. Li, and Y. Liu, "From heuristic optimization to dictionary learning: a review and comprehensive comparison of image denoising algorithms," *IEEE Trans. Cybern.*, vol. 44, no. 7, pp. 1001-1013, Jul. 2014.

[5] A. Buades, B. Coll, and J. M. Morel, "A nonlocal algorithm for image denoising," in *Proc. IEEE International Conf. on Computer Vision Pattern Recognition (CVPR)*, pp. 60-65, 2005.

[6] K. Dabov, A. Foi, V. Katkovnik, and K. Egiazarian, "Image denosing by sparse 3-D transform-domain collaborative filtering," *IEEE Trans. Image Process.*, vol. 16, no. 8, pp. 2080-2095, Aug. 2007.

[7] J. Mairal, F. Bach, J. Ponce, G. Sapiro, and A. Zisserman, "Non-local sparse models for image restoration," in *Proc. IEEE International Conf. on Computer Vision (ICCV)*, pp. 2272-2279, 2009.

[8] W. S. Dong, X. Li, D. Zhang, and G. Shi, "Sparsity-based image denoising via dictionary learning and structural clustering," in *Proc. IEEE Conf. Computer Vision and Pattern Recognition (CVPR)*, pp. 457-464, 2011.

[9] W. S. Dong, L. Zhang, G. M. Shi, and X. Li, "Nonlocally centralized sparse representation for image restoration," *IEEE Trans. Image Process.*, vol. 22, no. 4, pp. 1620-1630, Nov. 2013.

[10] M. Aharon, M. Elad, and A. Bruckstein, "K-SVD: an algorithm designing overcomplete dictionaries for sparse representation," *IEEE Trans. Signal Process.*, vol. 54, no. 11, pp. 4311-4322, Nov. 2006.

[11] M. Elad, and M. Aharon, "Image denoising via sparse and redundant representations over learned dictionaries," *IEEE Trans. Image Process.*, vol. 15, no. 12, pp. 3736-3745, Dec. 2006.

[12] D. Zoran, and Y. Weiss, "From learning models of natural image patches to whole image restoration," in *Proc. IEEE International Conf. on Computer Vision (ICCV)*, pp. 479-486, 2011.

[13] X. M. Liu, D. M. Zhai, D. B. Zhao, G. T. Zhai, and W. Gao, "Progressive image denoising through hybrid graph laplacian regularization: a unified framework,*" IEEE Trans. Image Process.*, vol. 23, no. 4, pp. 1491-1503, Apr. 2014.

[14] J. Xu, L. Zhang, W. M. Zuo, D. Zhang, and X. C. Feng, "Patch group based nonlocal self-similarity prior learning for image denoising," in *Proc. IEEE International Conf. on Computer Vision (ICCV)*, pp. 244-252, 2015.

[15] S. Sahoo, and A. Makur, "Enhancing image denoising by controlling noise incursion in learned dictionaries," *IEEE Signal Process. Lett.*, vol. 22, no. 8, pp. 1123-1126, Aug. 2015.

[16] X. H. Zeng, W. Bian, W. Liu, J. L. Shen, and D. C. Tao, "Dictionary pair learning on grassmann manifolds for image denosing," *IEEE Trans. Image Process.*, vol. 24, no. 11, pp. 4556-4569, Nov. 2015.

[17] B. B. Ni, P. Moulin, and S. C. Yan, "Order preserving sparse coding," *IEEE Trans. Patt. Anal. Mach. Intell.*, vol. 35, no. 2, pp. 367-380, Feb. 2013.

[18] A. Levin, B. Nadler, F. Durand, and W. T. Freeman, "Patch complexity, finite pixel correlations and optimal denoising," in *Proc. European Conf. on Computer Vision (ECCV)*, pp. 73-86, 2012.

[19] G. Shikkenawis, and S. K. Mitra, "2D orthogonal locality preserving projection for image denoising," *IEEE Trans. Image Process.*, vol. 25, no. 1, pp. 262-273, Jan. 2016.

[20] M. Zheng, J. Bu, C. Chen, C. Wang, L. Zhang, G. Qiu, and D. Cai, "Graph regularized sparse coding for image representation," *IEEE Trans. Image Process.*, vol. 20, no. 5, pp. 1327-1336, May 2011.

[21] A. Kheradmand, and P. Milanfar, "A general framework for regularized, similarity-based image restoration," *IEEE Trans. Image Process.*, vol. 23, no. 12, pp. 5136-5151, Dec. 2014.

[22] L. Zhuang, S. Gao, J. Tang, J. Wang, J. Lin, Y. Ma, and N. Yu, "Constructing a nonnegative low-rank and sparse graph with data-adaptive features," *IEEE Trans. Image Process.*, vol. 24, no. 11, pp. 3717-3728, Nov. 2015.

[23] J. Wang, J. Yang, K. Yu, F. Lv, T. Huang, and Y. Gong, "Locality-constrained linear coding for image classification," in *Proc. IEEE Conf. Computer Vision and Pattern Recognition (CVPR)*, pp. 3360-3367, 2010.

[24] S. Lu, Z. Wang, T. Mei, G. Guang, and D. Feng, "A bag-of-importance model with locality-constrained coding based feature learning for video summarizaiton," *IEEE Trans Multimedia*, vol. 16, no. 6, pp. 1497-1509, Oct. 2014.